\documentclass[conference]{IEEEtran}
\IEEEoverridecommandlockouts

\usepackage{cite}
\usepackage{amsmath,amssymb,amsfonts}
\usepackage{algorithmic}
\usepackage{graphicx}
\usepackage{textcomp}
\usepackage{xcolor}
\usepackage{tikz}
\usepackage{array}
\usepackage{url}
\usepackage{orcidlink}

\makeatletter
\newcommand{\linebreakand}{%
  \end{@IEEEauthorhalign}
  \hfill\mbox{}\par
  \mbox{}\hfill\begin{@IEEEauthorhalign}
}
\makeatother

\newcommand\copyrighttext{%
      \footnotesize \textcopyright 2023 IEEE. Personal use of this material is permitted.
      Permission from IEEE must be obtained for all other uses, in any current or future
      media, including reprinting/republishing this material for advertising or promotional
      purposes, creating new collective works, for resale or redistribution to servers or
      lists, or reuse of any copyrighted component of this work in other works.}
    \newcommand\copyrightnotice{%
    \begin{tikzpicture}[remember picture,overlay]
    \node[anchor=south,yshift=10pt] at (current page.south) {\fbox{\parbox{\dimexpr\textwidth-\fboxsep-\fboxrule\relax}{\copyrighttext}}};
    \end{tikzpicture}%
}

\def\BibTeX{{\rm B\kern-.05em{\sc i\kern-.025em b}\kern-.08em
    T\kern-.1667em\lower.7ex\hbox{E}\kern-.125emX}}
\begin{document}

\title{Deep learning-based stereo camera multi-video synchronization}
\author{\IEEEauthorblockN{1\textsuperscript{st} Nicolas Boizard}
\IEEEauthorblockA{\textit{ISIA Lab - University of Mons} \\
Mons, Belgium \\
nicolas.bzrd@gmail.com}
\and
\IEEEauthorblockN{2\textsuperscript{nd} Kevin El Haddad \orcidlink{0000-0003-1465-6273}\IEEEauthorrefmark{1}\IEEEauthorrefmark{2}}
\IEEEauthorblockA{\IEEEauthorrefmark{1}\textit{ISIA Lab - University of Mons} \\
\textit{\IEEEauthorrefmark{2}The Big Projects} \\
Mons, Belgium \\
kevin.elhaddad@umons.ac.be
}
\and
\IEEEauthorblockN{3\textsuperscript{rd} Thierry Ravet}
\IEEEauthorblockA{\textit{ISIA Lab - University of Mons} \\
Mons, Belgium \\
thierry.ravet@umons.ac.be}
\linebreakand
\IEEEauthorblockN{4\textsuperscript{th} François Cresson}
\IEEEauthorblockA{\textit{ISIA Lab - University of Mons} \\
Mons, Belgium \\
francois.cresson@umons.ac.be}
\and
\IEEEauthorblockN{5\textsuperscript{th} Thierry Dutoit}
\IEEEauthorblockA{\textit{ISIA Lab - University of Mons} \\
Mons, Belgium \\
thierry.dutoit@umons.ac.be}
}

\maketitle
\copyrightnotice

\begin{abstract}
Stereo vision is essential for many applications. Currently, the synchronization of the streams coming from two cameras is done using mostly hardware. A software-based synchronization method would reduce the cost, weight and size of the entire system and allow for more flexibility when building such systems. With this goal in mind, we present here a comparison of different deep learning-based systems and prove that some are efficient and generalizable enough for such a task. This study paves the way to a production ready software-based video synchronization system.
\end{abstract}


\section{INTRODUCTION}
\label{sec:intro}
Computer vision applications such as visual depth estimation, useful in the areas of autonomous navigation and human-agent interaction, require stereo vision. Currently, the most accurate stereo systems are based on specific hardware solutions such as stereo cameras. The synchronization of video streams within these cameras is therefore achieved using electronic systems that increase the cost, the weight, limit the flexibility of the systems and require more space. Replacing these electronic solutions with robust software would make these systems more flexible, allowing them to be used with different camera configurations and settings. Also, making these software solutions open-source would help reduce the costs to build stereo vision systems and thus make them more accessible to a wider community. In this article, we present a study on the automatic synchronisation in time of two video sequences from two different lenses/cameras filming a common area with a deep learning method.
Research for this specific application is surprisingly neglected.
Indeed previous work did tackle the problem of software-based multi-camera synchronization \cite{wireless, faizullin2020twist, actionbased, yin2022self}. But, the novelty and advantage of our work is that it relies solely on the image content (the pixels) and is not dependant on hardware components or their features (internal clocks, timestamps, network, etc.). It does not require prior knowledge to function nor is it limited to a specific application.
To the best of our knowledge, the closest related work on our previously defined problem to date is the work~\cite{journeyVideoSynchronization}. The objective of their research was to synchronize several travelling paths despite changing weather conditions. Based on the triplet loss method, their model was able to synchronize two video sequences with 38\% of the predicted frame pairs having a shift of less than four frames. Their objective is quite different from our problematic. Indeed, our problem focuses on the synchronization of videos acquired by two cameras with overlapping scenes captured over a similar time periods, as opposed to the alignment of videos recorded by a single camera following an identical trajectory but at different time periods, which involves serious changes in the scenes.

In this paper, we chose to divide the problem of video synchronization into two subproblems: given two video sequences of twenty frames captured at the same frame rate by two different cameras with a distance between their optical centres limited to a few centimeters and oriented towards the same direction, a first module computes one by one the correspondence scores between each frame of the two sequences; then, when all the correspondence scores are computed, a second module takes as input all these scores to estimate the average delay between the sequences. This final frame delay can then be used to synchronise the two sequences. The need for a second submodule in our system is due to the fact that the output of the first one does not give consistent results in terms of frame distance when assigning similar images: images A and B of sequence 1 could be assigned to images C and D of sequence 2 with a frame distance from A to C different from the one from B to D.

A commonly used method for the similarity estimation between a pair of images is to use the SIFT algorithm \cite{sift}. This algorithm allows to detect, describe and locate common features between two images. However, preliminary tests with it gave poor results, so we decided to focus our experiments, in this work, on deep learning-based systems which initially showed more promising results.
Siamese networks are another common choice for image pair comparison. The authors in~\cite{siameseImagePairs}, use such architecture and show that the features proposed using a Siamese network improve the matching performance compared to the features obtained with a classical CNN networks trained for image classification. Other works also based on Siamese networks \cite{imageMatching, faceSiamese, humanSiamese, signatureSiamese} as well as triplet networks \cite{humanTriplet, searchTriplet} showing encouraging results, motivated our choice towards considering them in our experiments. The main contribution of this work is a comparative study of different architectures and approaches, a product of which being what we believe to be a first deep learning-based stereo video time synchronization system. We thus prove the feasibility of a software based video synchronization. 
The source code and the dataset used here are provided to the community~\footnote{\url{https://github.com/numediart/multi_video_sync}}.

We therefore first present the dataset collected in Section~\ref{sec:dataset}. We then present the systems, experiments and results in Sections~\ref{sec:systems}, \ref{sec:experiments} and \ref{sec:results} before finally concluding in Section~\ref{sec:conclusion}.


\section{DATASET}
\label{sec:dataset}
\begin{table*}[ht]
\caption{List of the different datasets (image + optical flow), with their properties. These are: the camera used, the environment (2 possible locations), whether the camera is in motion or not, and whether the scenes contain moving objects (human walking). The columns starting with the abbreviation No. indicate the Number of image and optical flow pairs that the dataset contains as well as the number of video sequences formed on these two types of data. To differentiate a dataset according to the type of data chosen, "*" in the name will be replaced by "img" for images and "flw" for the optical flow in the rest of this paper.}
\begin{center}
\begin{tabular}{|m{1.25cm}|m{2.2cm}|m{1.3cm}|m{1.2cm}|m{1.1cm}|m{1.7cm}|m{1.8cm}|m{1.7cm}|m{1.7cm}|}
\hline
    Name    & Camera          & Location & Moving Camera & Moving Objects & No. of raw img & No. of optical flow img & No. of raw img seq & No. of optical flow seq  \\ \hline
*\_zed\_p   & Stereolab Zed 2 & Loc1     & Pushed        & No             & 23892          & 23890                   & 40000              & 40000\\ \hline
*\_zed\_p2  & Stereolab Zed 2 & Loc2     & Pushed        & No             & 16104          & 16102                   & 40000              & 40000\\ \hline
*\_zed\_m   & Stereolab Zed 2 & Loc1     & Still         & Yes            & 3960           & 3958                    & 3000               & 3000 \\ \hline
*\_zed\_pm  & Stereolab Zed 2 & Loc1     & Pushed        & Yes            & 4200           & 4198                    & 3000               & 3000 \\ \hline
*\_oak\_p   & Oak D Lite      & Loc1     & Pushed        & No             & 7910           & 7908                    & 3000               & 3000 \\ \hline
*\_oak\_m   & Oak D Lite      & Loc1     & Still         & Yes            & 5714           & 5712                    & 3000               & 3000 \\ \hline
*\_oak\_pm  & Oak D Lite      & Loc1     & Pushed        & Yes            & 4784           & 4782                    & 3000               & 3000 \\ \hline
\end{tabular}
\end{center}
\label{datasetsResume}
\end{table*}

For our experiments, matching/non-matching pairs of images and image sequences were needed. It is also important to have videos recorded in different configurations/environments to study the robustness of our models. We therefore recorded videos using stereo cameras at 30 fps in RGB mode using two different cameras  (Stereolab Zed 2\footnote{https://www.stereolabs.com/zed-2/} and OAK D Lite\footnote{https://shop.luxonis.com/products/oak-d-lite-1}) and in different environments. Other parameters considered were: the camera being in motion or fixed and the presence or not of moving objects in the scene (humans walking in our case). Note that recording videos with stereo cameras allowed us to desynchronize them manually by applying different delays. The results are summarized in Table~\ref{datasetsResume}.

The first module of our system mentioned in the introduction, the image matching model, required unmatched image pairs along with the matching pairs for the training. We therefore created matching and non-matching pairs of images based on the recordings. Matching pairs were formed by the corresponding images of the left and right lenses of the camera. The non-matching pairs were formed by randomly pairing images from the left and right lenses with the following conditions: they must come from the same set and the frame coming from the right must be located in an interval $t \in [-10,-1]$ or $t \in [1,10]$ with t the capture time of the left frame. We also ensure that all images in the sets are 224x224px in size and in grayscale by post-processing them.

Also, optical flow features were extracted using the Farneback algorithm \cite{farnebackOpticalFlow} from the videos recorded. Our raw image datasets being relatively small, we rely on the optical flow to provide additional motion information to our models in order to avoid overfitting in some cases. Indeed, as we show in the following sections, these features turned out to be more efficient than the raw images in several cases. This results in total of fourteen sets summarized in Table~\ref{datasetsResume} (seven of raw images and seven of optical flow data).

Concerning the second module mentioned in the introduction, the delay estimation module, we desynchronized two corresponding sequences of twenty images coming from the left and the right lenses of a stereo camera, by a random delay varying from zero to twenty frames. A delay of zero frames, corresponds to synchronized sequences while a delay of twenty frames corresponds to completely desynchronized ones (there are therefore no frames in common between the left and right videos). The delay between each pair of video clips is randomly assigned to ensure balanced datasets. The total number of sequences for each of the raw images and optical flow sequences can be seen in Table~\ref{datasetsResume}.


\section{SYSTEMS}
\label{sec:systems}
As mentioned in the introduction, our problem is divided into two subproblems: firstly, finding the matching frames between our videos and secondly, calculating the delay using these matches. Our model therefore consists of two submodels. The first submodule, called "Matching Frames", will compute a matching score between two frames. Using the calculation of these matching scores, we construct a matching matrix. Indeed, by calculating the score of the first frame of the left video with all the frames of the right video, then the score of the second frame of the left video with all the frames of the right video and so on, we obtain a matching matrix on all the frames of the videos. Finally, with this correspondence matrix, our second submodule called "Delay Estimation" estimates the delay that may exist between the two sequences.

\begin{figure}[ht]
\includegraphics[width=7cm]{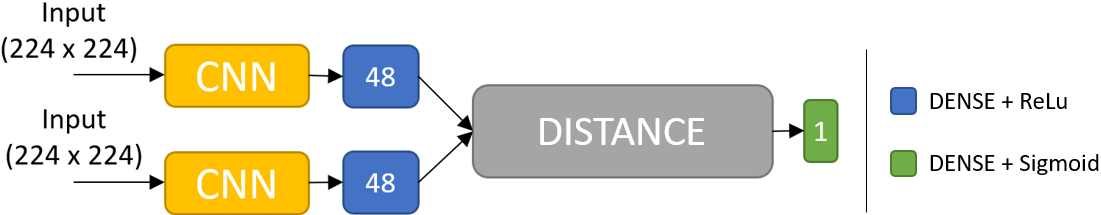}
\centering
\caption{Diagram of "CNNSiamese" model. The CNN blocks are summarised in the Table \ref{layersCNNSiamese} and the Distance block performs a subtraction.}
\label{siameseModel}
\end{figure}

For the "Matching Frames" submodule, several network architectures were considered. The first one is a Siamese network called "CNNSiamese" and described in the Fig \ref{siameseModel}. "CNNSiamese" is trained with the binary crossentropy loss and the Adam optimizer. As input, the network takes a frame from the left lens and another from the right lens. On the output, a neuron activated by a Sigmoid function gives a matching score between these two frames. The closer the score is to 1, the higher the network considers that the pair of frames match. The stride of the convolution layers is set to (1,1) and activated with the ReLu function.

Another version of "CNNSiamese" with CNN blocks based on a pre-trained EfficientNet \cite{efficientNet} architecture was also tested in this work. Due to poor results in the early stages of the experiments, this model was dropped from the study.

\begin{table}[ht]
\caption{Layers in the CNN blocks for "CNNSiamese", "TripletEuc" and "TripletSim" model.}
\begin{center}
\begin{tabular}{|l|l|l|}
\hline
Layers                 & Units/Filters & Kernel Size \\ \hline
Conv2D                 & 32            & (3,3)       \\
MaxPooling2D           & X             & (2,2)       \\
Conv2D                 & 48            & (3,3)       \\
MaxPooling2D           & X             & (2,2)       \\
Conv2D                 & 48            & (3,3)       \\
MaxPooling2D           & X             & (2,2)       \\
Conv2D                 & 64            & (3,3)       \\
MaxPooling2D           & X             & (2,2)       \\
GlobalAveragePooling2D & X             & X           \\ \hline
\end{tabular}
\end{center}
\label{layersCNNSiamese}
\end{table}

\begin{figure}[ht]
\includegraphics[width=6.5cm]{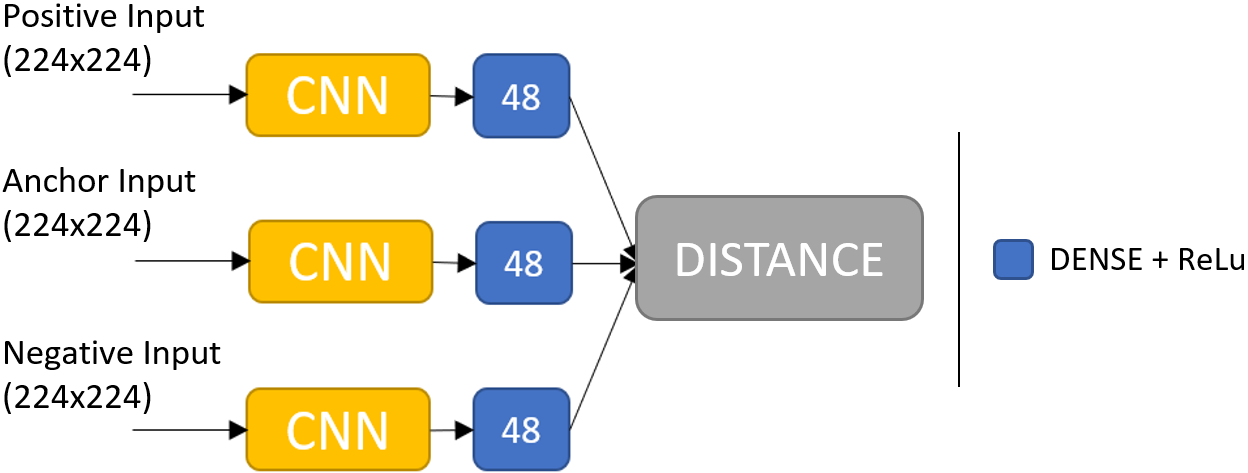}
\centering
\caption{Diagram of the triplet loss network. The layers used in the CNN blocks are summarised in the Table \ref{layersCNNSiamese}. The Distance block computes the distance between the positive/anchor pair and between the anchor/negative pair. The distance method is either the Euclidean distance or the Cosine similarity.}
\label{trippletModel}
\end{figure}

The second architecture considered and depicted in Figure~\ref{trippletModel} is based on the Triplet Loss method~\cite{tripletLoss}. A triplet loss architecture aims at minimizing the distance between two positively associated data and maximize the distance with two negatively associated data (the anchor associated with the positive and with negative input respectively as shown in Fig~\ref{trippletModel}). The distance obtained of either the positive/anchor or negative/anchor branch is then used here to fill the matching matrix forwarded to the second submodule. Two versions of the triplet loss network are designed. The first one, "TripletEuc" for which the distance is an Euclidean distance and whose training aims at minimizing the loss:\linebreak
\vspace{-3mm}
\begin{equation*}
     \resizebox{\hsize}{!}{$\sum_{i}^{N}\left[\left\|f\left(x_{i}^{a}\right)-f\left(x_{i}^{p}\right)\right\|_{2}^{2}-\left\|f\left(x_{i}^{a}\right)-f\left(x_{i}^{n}\right)\right\|_{2}^{2}+\alpha\right]$}
     \vspace{-1mm}
\end{equation*}
while for the second one, "TripletSim", the distance is the cosine similarity and trained to minimize the loss:
\vspace{-2mm}
\begin{equation*}
     \resizebox{\hsize}{!}{$\sum_{i}^{N}\left[\max(\cos(f\left(x_{i}^{a}\right),f\left(x_{i}^{p}\right))-\cos(f\left(x_{i}^{a}\right),f\left(x_{i}^{n}\right))+\alpha,0.0)\right]$}
     \vspace{-2mm}
\end{equation*}
with $N$ the number of samples per batch, $f\left(x\right)$ the output of a (CNN+Dense) branch, $x^{a}$ an anchor frame (frame of reference), $x^{p}$ a positive one (frame matching with reference), $x^{n}$ a negative one (frame non-matching with reference) and $\alpha$ the margin fixed at 0.5 for both formulas.

For the second submodule "Delay Estimation", two methods were compared. The first, labelled "HeatMap", consists in finding the diagonal of the similarity matrix containing the highest amount of maximum scores in order to calculate the distance between this diagonal and the main diagonal. This distance would represent the delay in number of frames since the main diagonal corresponds to a delay of zero. By nature this method is limited since it is more likely to choosing a diagonal with a higher number of values. The second method uses a dense neural network coupled with a categorical crossentropy loss. The network takes as input the correspondence matrix flattened and outputs forty neurons of a single dense layer activated with the Softmax function corresponding to a positive or negative delay between zero and twenty frames. The layers of this model named "DenseDelay" are detailed in the Table~\ref{sequentialDelayModel}.

\begin{table}[ht]
\caption{The layers making up "DenseDelay".}
\begin{center}
\begin{tabular}{| m{2cm} | m{1.8cm}| m{2.2cm}|}
\hline
Layers & Units & Activation \\ \hline
Input  & 400   & X          \\
Dense  & 64    & ReLu       \\
Dense  & 32    & ReLu       \\
Dense  & 40    & Softmax    \\ \hline
\end{tabular}
\end{center}
\label{sequentialDelayModel}
\end{table}

A CNN architecture taking the similarity matrix as input as well as a bidirectional LSTM network have also been tested. However, due to their poor performance, these two networks are not included in this study.


\section{EXPERIMENTS}
\label{sec:experiments}

In order to test the robustness of our models in different environments and with different cameras, several experiments were performed. They all aim at evaluating both the efficiency and the robustness of several systems built by combining different submodules of "Matching Frames" with different "Delay Estimation" submodules.
First we train the different possible models for the "Matching Frames" submodule seen at in section \ref{sec:systems}. All the trainings were performed on 100 epochs with a batch of 1 and a learning rate of 0.01 for the "CNNSiamese" networks and a batch of 32 with a learning rate of 0.001 for the "TripletEuc" and "TripletSim" networks. Each model were trained on 4 different datasets, "img\_zed\_p", "img\_zed\_p2", "flw\_zed\_p" and "flw\_zed\_p2". Training on these datasets allows to analyze the impact of the type of the data used (optical flow or raw images), evaluate the robustness on different cameras, and within different environments (see Table~\ref{datasetsResume}).
The trained first submodule is then combined, here with either the "DenseDelay" and "HeatMap" methods. The obtained systems are then evaluated based on the final delay estimation performances. As the "HeatMap" method is a simple comparison algorithm, it did not require any training. However, "DenseDelay" is trained each time on the data of the training set used to train the first submodule associated with it. These trainings are performed over 50 epochs with a learning rate of 0.01. The tests are then run with all possible combinations of first and second submodules, on all datasets that were not used for training.


\section{RESULTS}
\label{sec:results}

\begin{figure}[ht]
\includegraphics[width=6.5cm]{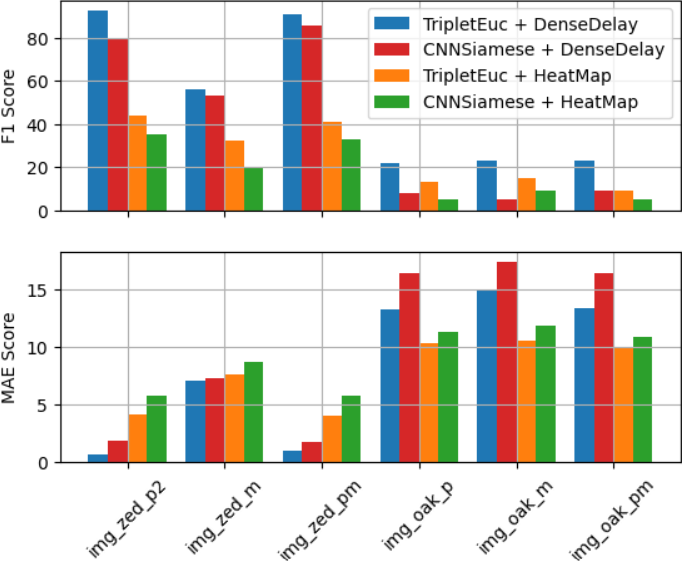}
\centering
\caption{F1 score \& MAE of different combination of submodules on different datasets (on the x-axis). The models were all trained with the "img\_zed\_p" dataset.}
\label{first_results}
\end{figure}

\begin{figure}[ht]
\includegraphics[width=6.5cm]{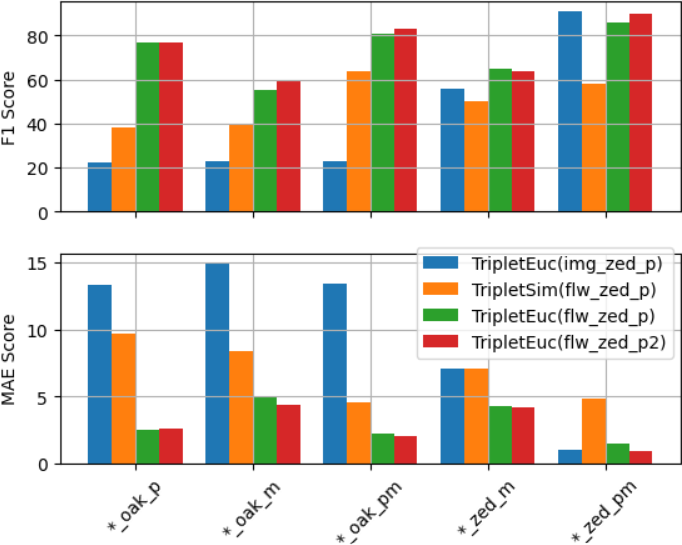}
\centering
\caption{F1 score \& MAE for systems with different first submodules and the same "DenseDelay" as the second submodule tested on different test sets (on the x-axis). The * in the names is to be replaced with the same type of data as the training set (in parentheses in the legend): image or optical flow.}
\label{second_results}
\end{figure}

All submodule combinations mentioned in Section~\ref{sec:experiments} were compared in our experiments. But for the shake of clarity, the results not bringing any valuable information were not reported here.
Figures~\ref{first_results} and \ref{second_results} show the F1 score estimating the amount of times the systems provided the exact delay, and Mean Absolute Error~(MAE) estimating the number of frames these systems were mistaken in, and this for different systems and on several datasets representing different acquisition contexts and data content.
The first and main conclusion from these results is that deep learning-based systems do have the potential to be used for video synchronization tasks. Indeed as can be seen in Fig.~\ref{second_results} the TripletEuc+DenseDelay when trained and used with optical flow data shows high F1 scores ranging between 60 and 90\% with an average of 74.8\%, low MAE across different datasets with a maximum average error of 4.14 frames with an average of 2.80 (ranging between 0.89 and 4.14).This shows both its effectiveness, generalizability and robustness on data, material and context. The second one is the use of optical flow data instead of raw images improves significantly the robustness of the system as can be seen when comparing both figures. To go further into detail, based on Fig.~\ref{first_results} we can see that the DenseDelay method for the delay estimation module outperforms the HeatMap one.


\section{CONCLUSION AND FUTURE WORK}
\label{sec:conclusion}

In this work, we showed that a stereo video synchronization solution using deep learning is feasible. We compared several systems and showed that some architectures perform and generalize well. Although further analyses are necessary to have a better understanding about the performance and generalization of the systems, the results presented here pave the way for the implementation of a production-level system in a near future.
In perspective, we will perform a more thorough study including more parameter variations, especially regarding the environments and recording setup. We will also consider the use of the transformer models \cite{transformers} that proved their efficiency in computer vision applications in general.
\vfill\pagebreak

\bibliographystyle{ieeetr}
\bibliography{refs}

\end{document}